\pdfoutput=1
\documentclass[11pt]{article}
\usepackage[preprint]{acl}
\usepackage{times}
\usepackage{latexsym}
\usepackage{enumitem}
\usepackage{ragged2e} 
\usepackage{graphicx}
\usepackage{array}
\usepackage{booktabs}
\usepackage{float}
\usepackage{multirow}
\usepackage{xcolor}

\usepackage{arydshln}
\usepackage[T1]{fontenc}
\usepackage[utf8]{inputenc}

\usepackage{listings}
\usepackage{algorithm}
\usepackage{algpseudocode}
\usepackage{amsmath}
\usepackage{microtype}
\usepackage{inconsolata}
\usepackage{graphicx}
\usepackage[para]{footmisc}

\title{\methodName: Enhancing Multi-Table Text-to-SQL Generation via Schema Simplification}

\author{Poojah Ganesan$^{1}$$^\textbf{*}$ \quad Rajat Aayush Jha$^{1}$$^\textbf{*}$ \quad Dan Roth$^{2}$ \quad Vivek Gupta$^{1}$\textsuperscript{\textdagger} \\
        $^{1}$Arizona State University $^{2}$University of Pennsylvania \\
        \texttt{\{pganesa4,rjha16,vgupt140\}}@asu.edu \quad \texttt{danroth}@seas.upenn.edu
        }

\usepackage{titlesec} 
\usepackage{listings} % For code examples
\usepackage{xcolor}   % For colored code formatting
\usepackage{titlesec} 

\lstdefinelanguage{Prompt}{
    basicstyle=\ttfamily\small,
    frame=single,
    backgroundcolor=\color{gray!10}, % Light gray background
    breaklines=true,                % Break lines for long texts
    keywordstyle=\color{blue}\bfseries,    % Keywords in bold blue
    stringstyle=\color{red},              % Strings in red
    commentstyle=\color{green!50!black},  % Comments in green
    morekeywords={CREATE, TABLE, SELECT, WHERE, AS, CASE, WHEN, THEN, ELSE, END, FROM, ORDER, BY, LIMIT},
    literate={\{}{{\textcolor{cyan}{\{}}}{1} % Open brace in cyan
             {\}}{{\textcolor{cyan}{\}}}}{1} % Close brace in cyan
             {self.table}{{\textcolor{magenta}{self.table}}}{10} % Highlight text inside braces
             {self.question}{{\textcolor{magenta}{self.question}}}{13} % Highlight text inside braces
             {self.description}{{\textcolor{magenta}{self.description}}}{16} % Highlight text inside braces
             {self.table.columns}{{\textcolor{magenta}{self.table.columns}}}{18} % Highlight text inside braces
             {df.column}{{\textcolor{magenta}{df.column}}}{9} % Highlight text inside braces
             {self.plan}{{\textcolor{magenta}{self.plan}}}{9} % Highlight text inside braces
             {self.plan}{{\textcolor{magenta}{self.plan}}}{9} % Highlight text inside braces
             {step.prompt}{{\textcolor{magenta}{step.prompt}}}{11} % Highlight text inside braces
             {self.name}{{\textcolor{magenta}{self.name}}}{9}
             {Example}{{\textbf{Example}}}{7}
             {MySQL_Code_Generation:}{{\textbf{MySQL Code Generation:}}}{22} % Bold for "MySQL Code Generation:"
             {Instructions:}{{\textbf{Instructions:}}}{12} % Bold for "Instructions:"
             {Step_1}{{\textbf{Step 1}}}{6}
             {Step_2}{{\textbf{Step 2}}}{6}
             {Step_3}{{\textbf{Step 3}}}{6}
             {Step_4}{{\textbf{Step 4}}}{6}
             {Step_5}{{\textbf{Step 5}}}{6}
             {Step_6}{{\textbf{Step 6}}}{6}
             {Question_:}{{\textbf{Question:}}}{9}
             {LLM_Step}{{\textbf{LLM Step}}}{8}                               % Bold "LLM Step"
             {SQL_Step}{{\textbf{SQL Step}}}{8}
             {Champion_}{{\textcolor{red}{'\%Champion\%'}}}{13} 
        {Win__}{{\textcolor{red}{'Win'}}}{5}
        {Round__}{{\textcolor{red}{'\%1st Round\%'}}}{14}
        {No_Win}{{\textcolor{red}{'No Win'}}}{8}
        {1936_}{{\textcolor{red}{'1936'}}}{6}
}

\lstdefinelanguage{SQL}{
  morekeywords={CREATE, TABLE, SELECT, CASE, WHEN, THEN, ELSE, END, IS, NULL, LIKE, SUBSTRING, CAST, AS, REGEXP_REPLACE},
  sensitive=true,
  keywordstyle=\color{blue}\bfseries,
  commentstyle=\color{green!60!black},
  stringstyle=\color{red},
  basicstyle=\ttfamily\footnotesize,
  moredelim=[is][\color{purple}]{`}{`}, % For column names or identifiersbackticks
  literate={CREATE}{{\color{blue}\textbf{CREATE}}}{5}
           {TABLE}{{\color{blue}\textbf{TABLE}}}{5}
           {SELECT}{{\color{blue}\textbf{SELECT}}}{6}
           {CASE}{{\color{blue}\textbf{CASE}}}{4}
           {WHEN}{{\color{blue}\textbf{WHEN}}}{4}
           {THEN}{{\color{blue}\textbf{THEN}}}{4}
           {ELSE}{{\color{blue}\textbf{ELSE}}}{4}
           {END}{{\color{blue}\textbf{END}}}{3}
           {IS}{{\color{blue}\textbf{IS}}}{2}
           {NULL}{{\color{purple}\textbf{NULL}}}{4}
           {LIKE}{{\color{purple}\textbf{LIKE}}}{4}
           {SUBSTRING}{{\color{purple}\textbf{SUBSTRING}}}{9}
           {CAST}{{\color{purple}\textbf{CAST}}}{4}
           {REGEXP_REPLACE}{{\color{purple}\textbf{REGEXP\_REPLACE}}}{13}
}

\usepackage{times}
\usepackage{latexsym}
\usepackage{float}
\usepackage[T1]{fontenc}
\usepackage[utf8]{inputenc}
\usepackage{microtype}
\usepackage{inconsolata}
\usepackage{xspace}  % Required for \xspace
\newcommand{\methodName}{{\sc UnJoin}\xspace}

\begin{document}
\maketitle
\begingroup\def\thefootnote{}\footnotetext{$\textbf{*}$These authors contributed equally to this work. \quad \quad \textdagger Primary superviser of this work.}\endgroup
\begin{abstract}

Recent advances in large language models (LLMs) have greatly improved Text-to-SQL performance for single-table queries. But, it remains challenging in multi-table databases due to complex schema and relational operations. Existing methods often struggle with retrieving the right tables and columns, generating accurate JOINs and UNIONs, and generalizing across diverse schemas. To address these issues, we introduce \methodName, a two-stage framework that decouples the retrieval of schema elements from SQL logic generation. 
In the first stage, we merge the column names of all tables in the database into a single-table representation by prefixing each column with its table name. This allows the model to focus purely on accurate retrieval without being distracted by the need to write complex SQL logic. In the second stage, the SQL query is generated on this simplified schema and mapped back to the original schema by reconstructing JOINs, UNIONs, and relational logic. Evaluations on SPIDER and BIRD datasets show that \methodName matches or exceeds the state-of-the-art baselines. \methodName uses only schema information, which does not require data access or fine-tuning, making it scalable and adaptable across databases. 
\end{abstract}

\section{Introduction}

\vspace{-0.5em}
Relational databases form the foundation for structured data management in domains such as finance, healthcare, and education. Accessing information from these databases typically requires writing SQL queries, a skill that demands technical expertise. Text-to-SQL, known as semantic parsing, addresses this barrier by translating natural language questions into executable SQL commands \cite{DBLP:journals/corr/cmp-lg-9503016, li2014constructing,li2023llmservedatabaseinterface}, allowing non-experts to interact with complex databases. 

\begin{figure}[t]
\centering
    \includegraphics[width=0.90\columnwidth]{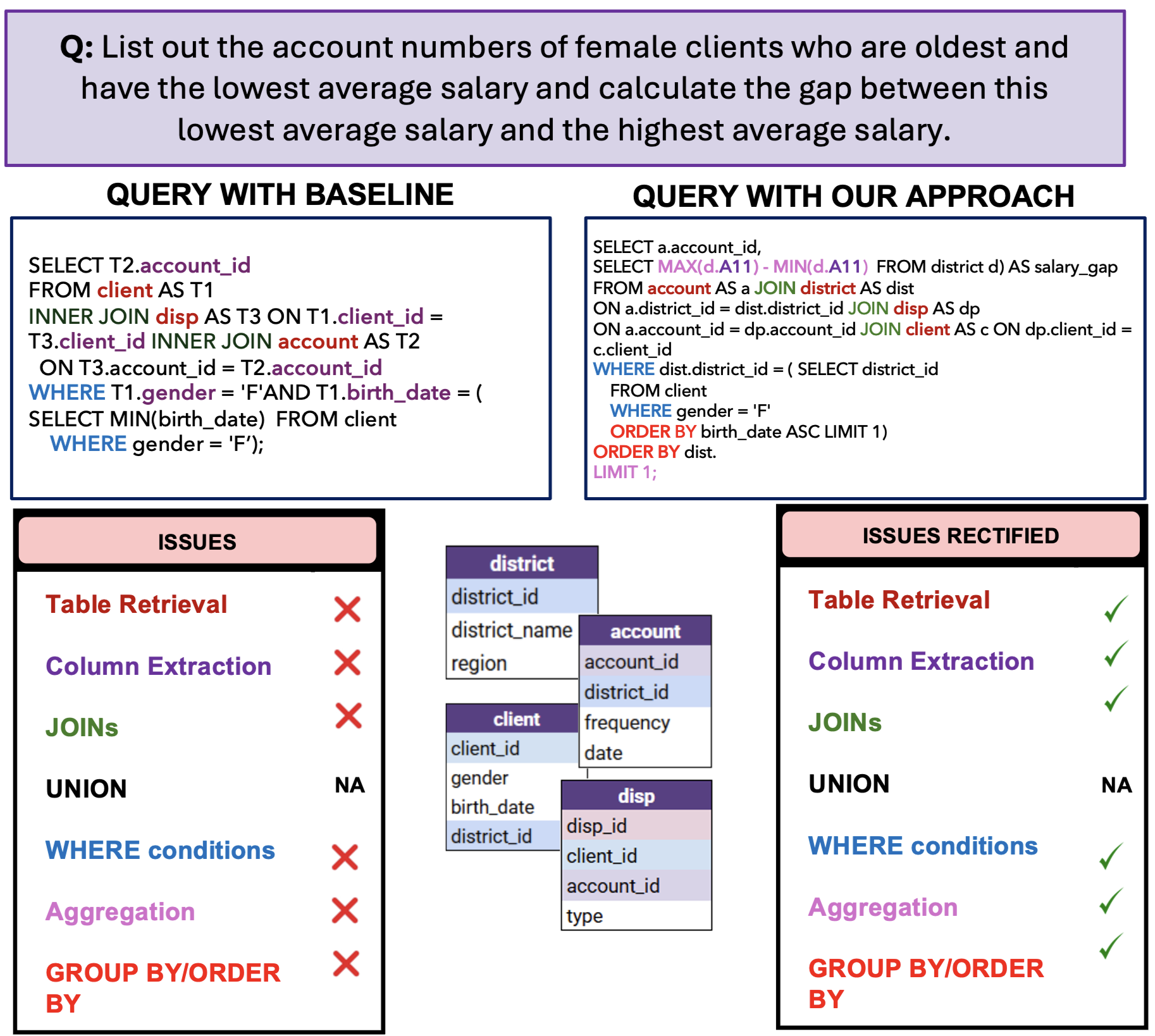}
    \vspace{-0.75em}
    \caption{Baseline vs \methodName}
    \vspace{-2.00em}
    \label{fig:challenges}
\end{figure}

 Early approaches use syntax trees or query sketches to guide query generation \cite{xu2017sqlnetgeneratingstructuredqueries, guo-etal-2019-towards}, while more recent methods rely on sequence-to-sequence models \cite{colin2020exploring, scholak-etal-2021-picard}. Recent advances leverage large language models (LLMs), either through in-context learning with powerful proprietary models or by fine-tuning smaller open-source alternatives, leading to substantial performance gains \cite{li2024codesbuildingopensourcelanguage, pourreza-rafiei-2024-dts}.   

While recent advances in Text-to-SQL have significantly improved performance on single-table databases, relatively little attention has been given to the more challenging multi-table setting. As shown in Figure~\ref{fig:challenges}, multi-table SQL generation introduces additional complexities such as identifying relevant tables and columns, resolving inter-table relationships (e.g., \texttt{JOIN}s and \texttt{UNION}s), and constructing more intricate queries involving \texttt{GROUP BY}, aggregation, ordering, and nested sub-queries. These challenges are central to building robust and generalizable Text-to-SQL systems for real-world applications. This raises a natural question: \emph{How can the progress made in single-table SQL generation be extended to handle the challenges of multi-table querying?}

To answer this, we propose \textbf{\methodName},  
a modular two-stage framework that decouples retrieval of schema elements from complex SQL generation. The key intuition is that LLMs are highly effective at generating SQL for single-table schemas, a setting they are more exposed to during training, while multi-table scenarios introduce structural complexity that is harder to handle directly. \methodName bridges this gap by reframing multi-table Text-to-SQL generation as a simplified single-table task that LLMs can solve more reliably. \textbf{\methodName} operates in two stages: 

(a.) \textbf{Stage 1: Schema Simplification:} The multi-table schema is flattened into a single-table format by merging the column names of all tables in the database into a single-table representation by prefixing each column with its table name, without altering the underlying data. 

(b.) \textbf{Stage 2: Query Generation and Translation.} A SQL query is first generated over this simplified schema and then translated back to align with the original schema by reconstructing necessary \texttt{JOIN}s, \texttt{UNION}s, and column relationships. By isolating the challenges of schema element selection and SQL logic construction, \textbf{\methodName} disambiguate the reasoning process and enables more accurate, scalable, and generalizable multi-table SQL generation. Our contributions are as follows:
\vspace{-0.5em}
\begin{enumerate}
    \item We propose \textbf{\methodName}, a novel approach for multi-table Text-to-SQL generation based on schema simplification. By decoupling table and column retrieval from complex SQL structuring, \methodName reduces compounding errors and improves robustness.
    \vspace{-0.5em}
    \item We demonstrate that our method outperforms a wide range of baselines, including (a) standard prompting, (b) in-context learning methods, (c) supervised fine-tuning approaches, (d) end-to-end table Question Answering (QA) models, and (e) recent reasoning-focused models such as Deepseek-R1 on both open book and closed book settings.
    \vspace{-0.5em}
    \item Through detailed analysis, we show that SQL generated by \methodName achieves superior performance in both table retrieval and column selection. Additionally, when combined with various retrievers in open-domain table QA settings, \methodName consistently outperforms standard few-shot prompting baselines.
    \vspace{-0.5em}
\end{enumerate}

\section{Methodology}

\begin{figure*}[t]
    \centering
    \includegraphics[width=0.70\textwidth]{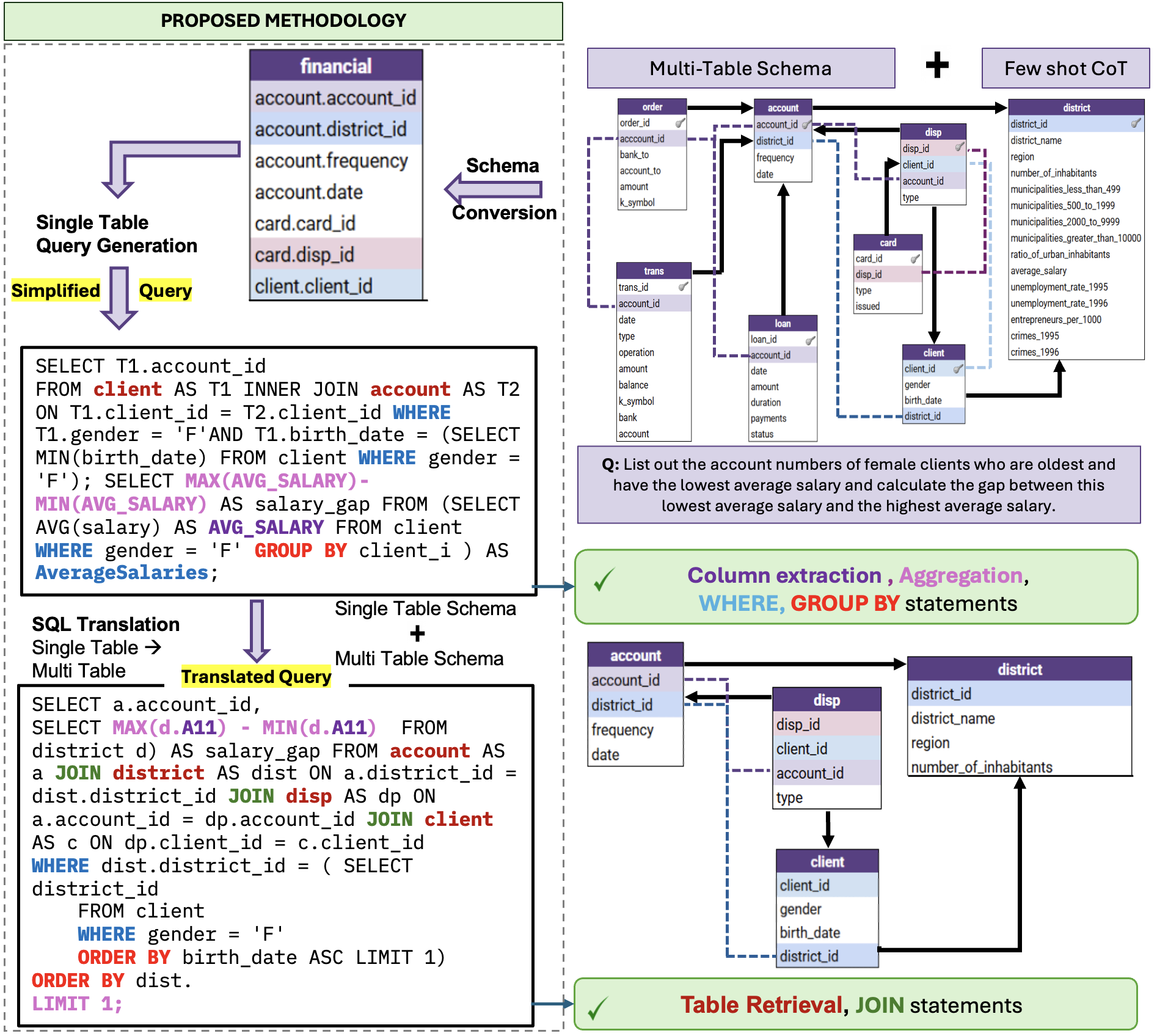}
    \vspace{-0.5em}
    \caption{\textbf{Our Proposed Method}. After schema conversion, the resulting single-table schema contains a total of 54 columns obtained by combining the columns from all shown tables: \textit{order, account, district, disp, card, client, loan,} and \textit{trans}. Due to space constraints, the complete single-table representation is not shown.
}\vspace{-1.75em}
    \label{fig:main_fig}
\end{figure*}

Existing LLM-based approaches struggle with identifying relevant tables and columns, resolving inter-table relationships like JOINs and UNIONs, and constructing more complex queries involving GROUP BY, nested subqueries, aggregation, and ordering. This is due to schema complexity, ambiguous column references, and limited context size. Overcoming these challenges is crucial for developing robust and generalizable Text-to-SQL systems suited for real-world applications. In Figure \ref{fig:main_fig}, we present our proposed framework, \textbf{\methodName}, which addresses these issues systematically through \textit{Schema Simplification} and \textit{Query Generation and Translation}. In the following section, a detailed
introduction of these steps are presented.
\vspace{-0.5em}
\paragraph{(1.) Schema Simplification}

To simplify complex multi-table schemas, we introduce a straightforward \textit{Schema Simplification} step. Here, we combine columns from all tables in the database into one simplified schema, without altering the underlying data. The goal is to create a single, flat table. We prefix each column name with its corresponding table name, which helps in removing ambiguity between similar column names from different table. 

Let's consider a database with six tables, each containing ten columns. The resulting single-table representation will comprise one table, named after the database, and a total of 60 columns. Here, each column name is reformatted using the structure \textit{TableName.ColumnName},  preserving its original context while eliminating ambiguity.
The algorithm for \textit{Schema Simplification} along with an example is shown in Appendix \ref{fig:schema-simplification} 

In the simplified format, the structure no longer depends on how the original tables were connected, so there is no concept of table joins at this stage. By removing the need to reason about table-to-table relationships, this step eliminates a layer of complexity for the LLM. The transformation relies only on schema-level information,  specifically, table and column names, and does not require access to row-level data. As a result, it generalizes well to databases of any size or content. This step is also fully deterministic and does not depend on LLMs, helping to reduce computational overhead and avoid hallucination errors that can occur when using LLMs.

\paragraph{(2.) Query Generation and Translation.}
This involves two sub-steps: \textit{\underline{(a.) Query Generation.}} and \textit{\underline{(b.) Query Translation.}}, which are described in detail below.

\noindent \textit{\underline{(a.) Query Generation.}} In this step, we use the \textit{Simplified Schema} obtained from the previous step to prompt the LLM to generate a semantically accurate intermediate (simplified) SQL query. By abstracting away relational complexities, such as JOIN and UNION operations, the LLM is free to focus solely on identifying relevant tables (which are represented as columns in this case of simplified schema), as well as accurately handling SQL operations like aggregation and numeric computations. This reduces the task to a simpler single-table Text-to-SQL scenario, a setting in which contemporary LLMs typically perform very well.

To ensure robust and unbiased query generation, the LLM is provided with a carefully structured prompt that includes the \textit{Simplified Schema}, the user’s question, detailed column descriptions and several schema-agnostic few-shot examples. These manually crafted examples remain consistent across evaluations to prevent data leakage or schema-specific biases. Importantly, as our approach never modifies the underlying table data, the resulting intermediate SQL is not directly executable; instead, its purpose is to clearly \textit{capture user intent within a simplified schema context}. The prompts are given in Appendix \ref{sec:prompts}

\textit{\underline{(b.) Query Translation.}}
The intermediate SQL query generated in the previous step references a simplified single-table schema and is therefore not directly executable. The \textit{Query Translation} step transforms this intermediate SQL into a fully executable SQL code, explicitly reintroducing necessary relational operations such as JOINs, pertaining to the original multi-table schema. \\
This process is fully automated, requiring only the original schema, the simplified schema, the simplified
SQL generated in the previous step, and the user question as input. Since this transformation is implicit within the LLM’s reasoning capabilities, no additional rule-based logic is required. To further reduce hallucination errors, we apply an edit-distance-based correction mechanism to ensure that the generated SQL aligns with the actual schema. This post-processing step only adjusts table and column names, ensuring that any abbreviations or modifications introduced by the LLM (e.g., table name \textit{disp} changed to \textit{disposition}) are accurately mapped back to their valid schema counterparts. In particular, this step does not alter SQL logic or introduce external constraints; it only enhances schema consistency. 
\vspace{-0.5em}
\paragraph{\methodName Varaints:} We explore two variations of our proposed method:
(a) \methodName$_{\text{SP}}$: Performs schema simplification, query generation, and translation sequentially within a single (joint) prompt approach.
(b) \methodName$_{\text{MP}}$: Separates these steps into distinct stages using a multi-prompt setup, allowing for more focused reasoning at each stage.

\vspace{-0.5em}
\section{Text2SQL Approaches}
\vspace{-0.5em}
We compare \methodName with five approches, as follows:
\vspace{-0.5em}
\paragraph{Standard Prompts:} We evaluate against prompting strategies like Direct Prompting with Few-Shot Chain of Thought (CoT) \cite{wei2023chainofthoughtpromptingelicitsreasoning}, Program of Thoughts (PoT) \cite{chen2023programthoughtspromptingdisentangling}, Meta Prompting (MP) \cite{suzgun2024metapromptingenhancinglanguagemodels}, and Self-Consistency (SC) \cite{wang2023selfconsistencyimproveschainthought}.\par\noindent

\vspace{-0.35em}
\paragraph{In-Context Learning (ICL):} We compare \methodName against several recent in-context learning baselines. DIN-SQL \cite{pourreza2023dinsqldecomposedincontextlearning} decomposes the task into subtasks, prompting GPT-4 separately for each. C3-SQL \cite{dong2023c3zeroshottexttosqlchatgpt} introduces schema filtering followed by calibrated prompting with self-consistency. RSL-SQL \cite{cao2024rslsqlrobustschemalinking} improves schema linking through bidirectional reasoning, contextual augmentation, and multi-turn self-correction. Several other recent in-context learning methods \cite{sheng2025basesqlpowerfulopensource, pourreza2024chasesqlmultipathreasoningpreference, lee-etal-2025-mcs} are not publicly available.

\vspace{-0.35em}
 \paragraph{Supervised Fine-Tuning (SFT):} Here, we compare against CodeS-7B \cite{li2024codesbuildingopensourcelanguage} and DTS-SQL \cite{pourreza-rafiei-2024-dts}.

\vspace{-0.35em}
\paragraph{End2End Table QA (TQA):} We also evaluate \methodName on End-to-End QA tasks. DATER \cite{ye2023largelanguagemodelsversatile}, TabSQLify \cite{nahid2024tabsqlifyenhancingreasoningcapabilities}, and ReActTable \cite{zhang2024reactable} decompose the table by generating SQL, typically using sequence-to-sequence architectures or fine-tuned LLMs, and then perform QA. We also compare against non-SQL based QA models such as MultiTabQA \cite{pal2023multitabqa} (Seq2Seq model), ResdSQL \cite{li2023resdsqldecouplingschemalinking} and two variats of QFMTS \cite{zhang2024qfmts}: QFMTS\_1 (summarization followed by QA), and QFMTS\_2 (single-table schema summarization followed by QA).\par\noindent

\paragraph{Reasoning LLMs:} This category consists of models optimized for complex reasoning tasks, including DeepSeek-R1-Distill-Qwen-7B \cite{deepseekai2025deepseekr1incentivizingreasoningcapability}, DeepSeek-R1-Distill-Llama-70B \cite{deepseekai2025deepseekr1incentivizingreasoningcapability}. \par\noindent

\vspace{-0.5em}
\section{Experimental Evaluation.}
\vspace{-0.5em}
\paragraph{Datasets:} We evaluate our approach on two widely used large-scale cross-domain Text-to-SQL datasets: Spider \cite{yu-etal-2018-spider} and BIRD \cite{li2023llmservedatabaseinterface}. Spider includes 200 databases, while BIRD contains 96, both spanning a diverse range of domains. In each dataset, tables are grouped by topic, with each topic corresponding to a separate database that includes an average of 5.4 tables, along with queries answerable using that schema.

\noindent Since our focus is on queries that require reasoning over multiple tables, we exclude those that involve only a single table, following a filtering strategy similar to \cite{chen2025tableretrievalsolvedproblem}. After filtering, we obtain 443 queries across 81 databases for Spider, and 1095 queries across 77 databases for BIRD.

\paragraph{Metrics:} A generated SQL query may be structurally valid and executable, yet still return an incorrect answer. To evaluate both correctness and execution, we use two key metrics: \textit{(a.) \underline{Query Execution Accuracy (QE)}} measures the percentage of generated queries that execute successfully without runtime errors. It checks only whether the query runs, not whether the result is correct, \textit{(b.) \underline{Exact Match (EM)}} captures the percentage of generated queries that not only execute successfully but also return the correct answer. Since EM requires both successful execution and correct output, it is a stricter metric and always a subset of QE. A query may be valid and count toward QE, yet yield an incorrect result, leading to a lower EM.

\paragraph{LLM's:} We evaluate the performance of Prompting based and Table-QA baselines (SQL-based) across three language models: GPT-4o \cite{openai2024gpt4ocard} (\textit{gpt-4o-2024-08-06}), Gemini 1.5 Flash \cite{geminiteam2024gemini15unlockingmultimodal}, and Llama 3.3 (70B)\footnote{https://github.com/meta-llama/llama-models}. Table-QA baselines (non SQL-based), ICL and SFT baselines are evaluated using GPT-4o. For further analysis about table and column retrieval and on variations of our approach baseline, we expand our evaluation to include additional models: GPT-4o Mini\footnote{https://openai.com/index/gpt-4o-system-card/} (\textit{gpt-4o-mini-2024-07-18}), Llama 3.1 (3B), SQLCoder (34B), Mixtral 7x8B \cite{jiang2024mixtralexperts} and CodeLlama \cite{roziere2024codellama}. This broader evaluation provides insight into the impact of model size and architecture on schema-aware SQL generation.

\paragraph{Evaluation Settings.} We evaluate our approach under two distinct settings: (a) Closed-book setting — The relevant database is provided in advance. Our method, \methodName, directly generates the final SQL query over the given multi-table schema, and (b) Open-book setting — Relevant tables must first be retrieved using state-of-the-art retrievers. Then, \methodName is applied to generate the final SQL query over the retrieved tables.

In the closed-book setting, the full database schema is available, enabling SQL generation without retrieval. In contrast, the open-book setting requires a retrieval step, where tables are selected based on both table-query similarity and table-table (i.e., joinability) similarity. This setting is more challenging because (i) the retrieval process may miss relevant tables (i.e., recall is less than 100$\%)$, and (ii) the retrieved tables may come from different databases, leading to invalid or spurious joins.

% \section{Results and Analysis}

\subsection{Results: Closed Book Settings}

\paragraph{Standard Prompts vs \methodName:}
As shown in Table \ref{tab:spider_baselines} , prompt-based baselines achieve high QE (e.g., CoT with 98.87 QE on SPIDER) but fall short in EM (75.28 EM), reflecting errors in table and column selection. In contrast, \methodName (\methodName$_{\text{MP}}$) maintains a strong QE (99.9) while significantly improving EM (77.13), demonstrating better schema understanding. A similar trend appears on BIRD (See Appendix \ref{tab:bird_baselines}), where \methodName$_{\text{MP}}$ surpasses the best baseline in EM (50.36 vs. 44.38).

\begin{table}[H]
\vspace{-0.75em}
\small
\centering
\footnotesize
\renewcommand{\arraystretch}{0.9}
\resizebox{\linewidth}{!}{%
\setlength{\tabcolsep}{3.8pt} % Adjust column spacing
\begin{tabular}{l|cccccc}
\hline
& \multicolumn{2}{c}{\textbf{GPT-4o}} & \multicolumn{2}{c}{\textbf{Gemini}} & \multicolumn{2}{c}{\textbf{LLAMA 70B}} \\
\textbf{Method} & \textbf{QE} & \textbf{EM} & \textbf{QE} & \textbf{EM} & \textbf{QE} & \textbf{EM} \\
\hline
\addlinespace[2pt]
& \multicolumn{6}{c}{\textit{Standard Prompts}} \\
\hline
\addlinespace[2pt]
CoT & 93.70 & 63.10 & 95.25 & 65.23 & 98.87 & 75.28 \\
PoT       & 93.77 & 67.39 & 94.40 & 72.89 & 96.71 & 68.61 \\
MP        & 94.14 & 67.76 & 92.91 & 67.39 & 90.88 & 63.86 \\
SC        & 94.14 & 67.76 & 94.48 & 72.89 & 96.35 & 67.21 \\
\hline
\addlinespace[2pt]
& \multicolumn{6}{c}{\textit{Table QA (SQL-based)}} \\
\hline
\addlinespace[2pt]
DATER       & 19.30 & 07.80 & 20.10 & 08.90 & 17.77 & 06.57 \\
TabSQLify   & 93.38 & 64.10 & 90.07 & 69.96 & 93.80 & 68.61 \\
ReActTable  & 02.00 & 00.20 & 03.00 & 00.20 & 02.80 & 00.30 \\
\hline
\addlinespace[2pt]
% \multicolumn{7}{c}{\textit{\methodName}} \\
% \hline
\addlinespace[2pt]
\methodName$_{\text{SP}}$ & \textbf{96.35} & \textbf{76.13} & 94.89 & 73.36 & 94.49 & 69.62 \\
\methodName$_{\text{MP}}$ & 94.89 & 76.00 & \textbf{95.99} & \textbf{75.57} & \textbf{99.90} & \textbf{77.13} \\
\hline
\end{tabular}
}
\vspace{-0.75em}
\caption{QE and EM scores on SPIDER dataset.}
\vspace{-0.75em}
\label{tab:spider_baselines}

\end{table}

\paragraph{ICL vs \methodName}: The results in Table \ref{tab:icl_baselines} show that RSL-SQL achieves the highest overall performance on both SPIDER and BIRD, particularly with strong generalization to BIRD (QE: 90.8, EM: 54.3). However, \methodName performs competitively, on SPIDER, where \methodName$_{\text{SP}}$ achieves an EM of 76.13, slightly surpassing RSL-SQL. On BIRD, both variants of \methodName outperform DIN-SQL and C3, showing robust cross-domain performance. 

\begin{table}[H]
\vspace{-0.5em}
\small
\centering
\setlength{\tabcolsep}{5pt}
\small
\begin{tabular}{l|cc|cc}
\hline
\multirow{2}{*}{\textbf{ICL Text-to-SQL}} & \multicolumn{2}{c|}{\textbf{SPIDER}} & \multicolumn{2}{c}{\textbf{BIRD}} \\
% \cline{2-5}
& \textbf{QE} & \textbf{EM} & \textbf{QE} & \textbf{EM} \\
\hline
DIN-SQL + GPT-4o   & 95.97 & 75.56 & 87.53  & 49.32 \\
C3 + GPT-4o       & 94.50 & 71.42 & - & - \\
RSL-SQL & \textbf{96.40} & 76.04 & \textbf{90.80} & \textbf{54.30} \\
\hline
\methodName$_{\text{SP}}$ (GPT-4o) & 96.35 & \textbf{76.13} & 88.55 & 51.74 \\
\methodName$_{\text{MP}}$ (GPT-4o) & 94.89 & 76.00 & 89.75 & 50.36 \\
\hline
\end{tabular}
% }
\vspace{-0.75em}
\caption{ICL models vs \methodName.}
\vspace{-1.5em}
\label{tab:icl_baselines}
\end{table}

\paragraph{End-to-end Table QA vs \methodName:} 

From Tables \ref{tab:spider_baselines} and \ref{tab:multi_table_em} it can be seen that \methodName outperforms all table QA baselines in both QE and EM by a substantial margin. This proves its efficiency and usefullness in end-to-end table QA tasks, apart from Text-to-SQL. It also scales efficiently on larger databases like BIRD, where other baselines struggle (see Appendix \ref{tab:bird_baselines}). MultiTabQA’s near-zero EM (0.03) suggests possible overfitting to single-table queries in SPIDER. We did not evaluate multi-table baselines on BIRD due to its large dataset size exceeding LLM input limits and the high cost of fine-tuning these models on larger datasets. As highlighted earlier, fine-tuned models often suffer from limited generalizability and become overly domain-specific, increasing costs while reducing their applicability to diverse scenarios. For these reasons, we restricted our experimentation to the SPIDER dataset. 

\begin{table}[!htbp]
\vspace{-0.5em}
\small
\centering
\setlength{\tabcolsep}{5pt}
\begin{tabular}{l|c}
\hline
\textbf{Table QA (non SQL-based) baselines} & \textbf{EM} \\
\hline
MultiTabQA (Seq2Seq) & 00.03 \\
QFMTS\_1 (GPT-4o)    & 25.28 \\
QFMTS\_2 (GPT-4o)    & 25.55 \\
ResdSQL              & 28.87 \\
\hline
\methodName$_{\text{SP}}$ (GPT-4o)    & \textbf{76.13} \\
\methodName$_{\text{MP}}$ (GPT-4o)    & 76.00 \\
\hline
\end{tabular}
\vspace{-0.75em}
\caption{Performance comparison on SPIDER dataset.}
\label{tab:multi_table_em}
\vspace{-1.0em}
\end{table}

\vspace{-1.0em}
\paragraph{SFT vs \methodName}: Across both datasets, \methodName demonstrates strong performance (Table \ref{tab:sft_baselines}), outperforming existing SFT baselines on SPIDER, and showing notable generalization on BIRD, highlighting the effectiveness of our schema simplification and decomposition strategy in improving multi-table Text-to-SQL generation.

\begin{table}[H]
\vspace{-0.5em}
\small
\centering
\setlength{\tabcolsep}{5pt}
\small
\begin{tabular}{l|cc|cc}
\hline
\multirow{2}{*}{\textbf{SFT Text-to-SQL}} & \multicolumn{2}{c|}{\textbf{SPIDER}} & \multicolumn{2}{c}{\textbf{BIRD}} \\
% \cline{2-5}
& \textbf{QE} & \textbf{EM} & \textbf{QE} & \textbf{EM} \\
\hline
CodeS-7B  & 94.14 &74.72& 89.65& 51.14 \\
DTS-SQL$_{
\text{DeepSeek 7B}}$ & 87.50 & 69.23& - & - \\
\hline
\methodName$_{\text{SP}}$ (GPT-4o) & \textbf{96.35 }& \textbf{76.13} & 88.55 & \textbf{51.74} \\
\methodName$_{\text{MP}}$ (GPT-4o) & 94.89 & 76.00 & \textbf{89.75} & 50.36 \\
\hline
\end{tabular}
% }
\vspace{-1em}
\caption{SFT models vs \methodName.}
\vspace{-1em}
\label{tab:sft_baselines}
\end{table}

\paragraph{Reasoning LLM's vs \methodName:} The results in Table \ref{tab:reasoning_models} highlight the strong performance of \methodName compared to reasoning model baselines in both QE and EM, demonstrating the versatility of our approach across different datasets.  

\begin{table}[H]
\vspace{-0.5em}
\small
\centering
\setlength{\tabcolsep}{5pt}
\small
\begin{tabular}{l|cc|cc}
\hline
\multirow{2}{*}{\textbf{Reasoning LM}} & \multicolumn{2}{c|}{\textbf{SPIDER}} & \multicolumn{2}{c}{\textbf{BIRD}} \\
% \cline{2-5}
& \textbf{QE} & \textbf{EM} & \textbf{QE} & \textbf{EM} \\
\hline
DeepSeek Qwen 7B & 91.21 & 57.82 & 66.25 & 25.01\\
DeepSeek Llama 70B & 95.26 & 61.08 & 84.12 & 44.54 \\
\hline
\methodName$_{\text{SP}}$ (GPT-4o) & \textbf{96.35 }& \textbf{76.13} & 88.55 & \textbf{51.74} \\
\methodName$_{\text{MP}}$ (GPT-4o) & 94.89 & 76.00 & \textbf{89.75} & 50.36 \\
\hline
\end{tabular}
% }
\vspace{-1em}
\caption{Reasoning models vs \methodName.}
\label{tab:reasoning_models}
\vspace{-0.75em}
\end{table}

\vspace{-1.5em}
\subsection{Results: Open Book Settings}

Table~\ref{tab:em_single_column_unjoin} presents the impact of \methodName on the EM accuracy of various retriever-based models. Across all retrievers—ranging from Contriever\footnote{4
https://huggingface.co/facebook/contriever-msmarco} and DTR \cite{herzig-etal-2021-open} to their enhanced counterparts in the JAR \cite{chen2025tableretrievalsolvedproblem} framework—\methodName consistently yields substantial improvements in SQL generation performance. Notably, the EM scores increase by 14.90 \% to 19.57 \%, demonstrating the robustness and generalizability of our approach.

\begin{table}[h]
\vspace{-0.75em}
\small
\centering
\begin{tabular}{lc}
\toprule
\textbf{Retriever} & \textbf{EM(\%): CoT $\rightarrow$ \methodName} ($\Delta$) \\
\midrule
ARM                 & 31.7 $\rightarrow$ 51.27 \textbf{(+19.57)} \\
Contriever  & 29.7 $\rightarrow$ 47.99 \textbf{(+18.29)} \\
DTR & 30.4 $\rightarrow$ 49.85 \textbf{(+19.45)} \\
JAR (DTR)                & 36.9 $\rightarrow$ 52.63 \textbf{(+15.73)} \\
JAR (Contriever)         & 36.2 $\rightarrow$ 51.10 \textbf{(+14.90)} \\
\bottomrule
\end{tabular}
\vspace{-0.75em}
\caption{Exact Match (EM) scores with \methodName in open book settings. $\Delta$ in parentheses shows the improvement.}
\label{tab:em_single_column_unjoin}
\vspace{-0.75em}
\end{table}

\vspace{-1.0em}
\paragraph{End2End Table Extraction.} Table~\ref{tab:retriever_UnJoin_all} presents a detailed comparison of multiple retriever models evaluated on precision (P) and recall (R) metrics, with the integration of our proposed \methodName framework for End2End table extraction settings. The results indicate a consistent and substantial improvement in retrieval performance across all retrievers when augmented with \methodName.

Specifically, both the DTR and Contriever retrievers exhibit significant performance gains, improving their precision by approximately 25\% and 27\%, and their recall by around 22\% and 26\% respectively. Even retrievers with relatively strong baseline performance, such as JAR (DTR) and JAR (Contriever), benefit  from \methodName, achieving gains in precision and recall ranging from 4.6\% to 6.7\%.

\begin{table*}[!htbp]
\centering
\small
\begin{tabular}{lcc|cc|cc|cc|cc}
\toprule
\bf Retriever & \multicolumn{2}{c|}{\textbf{ARM}} 
& \multicolumn{2}{c|}{\textbf{JAR (DTR)}} 
& \multicolumn{2}{c|}{\textbf{JAR (Contriever)}} 
& \multicolumn{2}{c|}{\textbf{DTR}} 
& \multicolumn{2}{c}{\textbf{Contriever}} \\
& \textbf{P} & \textbf{R} 
& \textbf{P} & \textbf{R} 
& \textbf{P} & \textbf{R} 
& \textbf{P} & \textbf{R} 
& \textbf{P} & \textbf{R} \\
\midrule
\textbf{CoT}
& 32.88 & 74.47
& 65.60 & 66.10 
& 64.70 & 65.40 
& 59.50 & 59.20 
& 55.90 & 55.50 \\
\textbf{\methodName}  
& 84.30 & 83.20
& 72.31 & 70.79 
& 70.84 & 70.00 
& 84.55 & 81.75 
& 82.97 & 81.88 \\
\midrule
\textbf{$\Delta$} 
& \textbf{+51.42} & \textbf{+08.73} 
& \textbf{+06.71} & \textbf{+04.69} 
& \textbf{+06.14} & \textbf{+04.60} 
& \textbf{+25.05} & \textbf{+22.55} 
& \textbf{+27.07} & \textbf{+26.38} \\
\bottomrule
\end{tabular}
\vspace{-0.75em}
\caption{Precision (P) and Recall (R) for all retriever configurations with \methodName for open book settings. $\Delta$ indicates the absolute improvement.}
\label{tab:retriever_UnJoin_all}
\vspace{-0.75em}
\end{table*}

These results highlight the core strength of our \methodName method—its ability to consistently enhance SQL generation across a range of retrievers. By simplifying schemas and structuring query decomposition, \methodName acts as a plug-in module that improves both base retrievers (e.g., Contriever, DTR) and advanced systems like JAR and ARM \cite{chen2025retrieveoncearmalignmentoriented}. This demonstrates its broad applicability and effectiveness in handling multi-table Text-to-SQL challenges.

\vspace{-0.35em}
\section{Discussion}
\vspace{-0.35em}
In this section, we further present ablation studies and additional insights from our \methodName approach for the closed book settings. 

\paragraph{How does \methodName benefit End2End table extraction?}

We analyze End2End table extraction performance where we asses how \methodName approach enhances table retrieval and column extraction, along with its impact on query accuracy. 

\begin{table*}[t]
\centering
\scriptsize
\renewcommand{\arraystretch}{1.0}
\setlength{\tabcolsep}{4pt}
\resizebox{\textwidth}{!}{%
\begin{tabular}{l|cccc|cccc}
\hline
\textbf{Model} 
& \textbf{CoT} & \textbf{CoT-SS} & \textbf{\methodName$_{\text{SP}}$} & \textbf{\methodName$_{\text{MP}}$} 
& \textbf{CoT} & \textbf{CoT-SS} & \textbf{\methodName$_{\text{SP}}$} & \textbf{\methodName$_{\text{MP}}$} \\
% \cline{2-9}
& \multicolumn{4}{c|}{\textbf{Table Precision (\%)}} 
& \multicolumn{4}{c}{\textbf{Table Recall (\%)}} \\
\hline
GPT-4o           & 59.30 & 60.90 & \textbf{84.10} & 83.00 & 58.90 & 61.50 & \textbf{86.70} & 84.10 \\
GPT-4o Mini      & 64.00 & 80.00 & 96.50 & \textbf{96.60} & 63.00 & 80.00 & 97.60 & \textbf{98.20} \\
Gemini 1.5 Flash & 55.10 & 76.90 & 72.90 & \textbf{78.10} & 54.80 & 74.40 & 75.90 & \textbf{78.60} \\
Mixtral          & 45.90 & 53.50 & \textbf{71.70} & 71.00 & 40.50 & 53.30 & \textbf{78.30} & 74.70 \\
SQLCoder-7B      & 52.90 & 64.00 & 48.40 & \textbf{65.00} & 53.60 & 69.30 & 53.00 & \textbf{70.20} \\
SQLCoder-34B     & 54.80 & 67.50 & \textbf{68.10} & 58.10 & 55.40 & 67.10 & \textbf{73.90} & 62.00 \\
Llama 3.1 (3B)   & 55.40 & \textbf{65.90} & 52.90 & 49.70 & 56.00 & \textbf{65.80} & 62.00 & 48.30 \\
Llama 3.3 (70B)  & 40.80 & 74.90 & \textbf{77.10} & 72.70 & 41.20 & 74.00 & \textbf{79.10} & 75.50 \\
% \hline
& \multicolumn{4}{c|}{\textbf{Column Precision (\%)}} 
& \multicolumn{4}{c}{\textbf{Column Recall (\%)}} \\
\hline
GPT-4o           & 50.70 & 67.40 & 88.10 & \textbf{88.60} & 53.70 & 71.10 & \textbf{95.50} & 95.30 \\
GPT-4o Mini      & 46.70 & 71.00 & 87.70 & \textbf{87.70} & 49.80 & 75.20 & 94.60 & \textbf{95.40} \\
Gemini 1.5 Flash & 53.90 & \textbf{90.40} & 88.60 & 89.10 & 55.60 & 86.10 & \textbf{96.20} & 93.20 \\
Mixtral          & 40.00 & 65.80 & 80.10 & \textbf{87.40} & 37.40 & 57.00 & 89.50 & \textbf{93.00} \\
SQLCoder-7B      & 50.00 & 63.30 & 43.60 & \textbf{75.00} & 54.10 & 70.90 & 54.00 & \textbf{83.50} \\
SQLCoder-34B     & 45.20 & 70.70 & \textbf{81.20} & 75.40 & 48.80 & 60.00 & \textbf{93.20} & 77.00 \\
Llama 3.1 (3B)   & 50.60 & \textbf{58.90} & 43.40 & 52.00 & 54.70 & \textbf{58.90} & 66.70 & 54.30 \\
Llama 3.3 (70B)  & 38.90 & 84.50 & 86.70 & \textbf{87.50} & 42.60 & 89.60 & \textbf{96.00} & 96.60 \\
\hline
\end{tabular}
} % end of resizebox
\vspace{-0.75em}
\caption{Evaluation of Table/Column Precision and Recall (in \%) on \textbf{SPIDER} dataset. For BIRD Dataset results, please refer to Appendix \ref{sec:PR_BIRD}.}
\label{tab:table_column_precision_recall_spider}
\vspace{-1.5em}
\end{table*}

Table~\ref{tab:table_column_precision_recall_spider} presents the precision and recall scores for table and column selection across different methods: standard CoT prompting, two variants of our approach—\methodName$_{\text{SP}}$ and \methodName$_{\text{MP}}$. Additionally, we introduce a new baseline, \underline{CoT-SS} (CoT on Simplified Schema), which directly generates the final multi-table SQL (including JOINs, etc.) from the intermediate simplified schema using few-shot chain-of-thought prompting. This comparison highlights the effectiveness of decoupling schema reasoning from SQL generation. Furthermore, we observe that \methodName$_{\text{SP}}$ and \methodName$_{\text{MP}}$ consistently outperform  other baselines (CoT, CoT-SS) in end-to-end table and column extraction.

\begin{figure}[!htbp]
\vspace{-0.75em}
    \centering
    \includegraphics[width=1\linewidth]{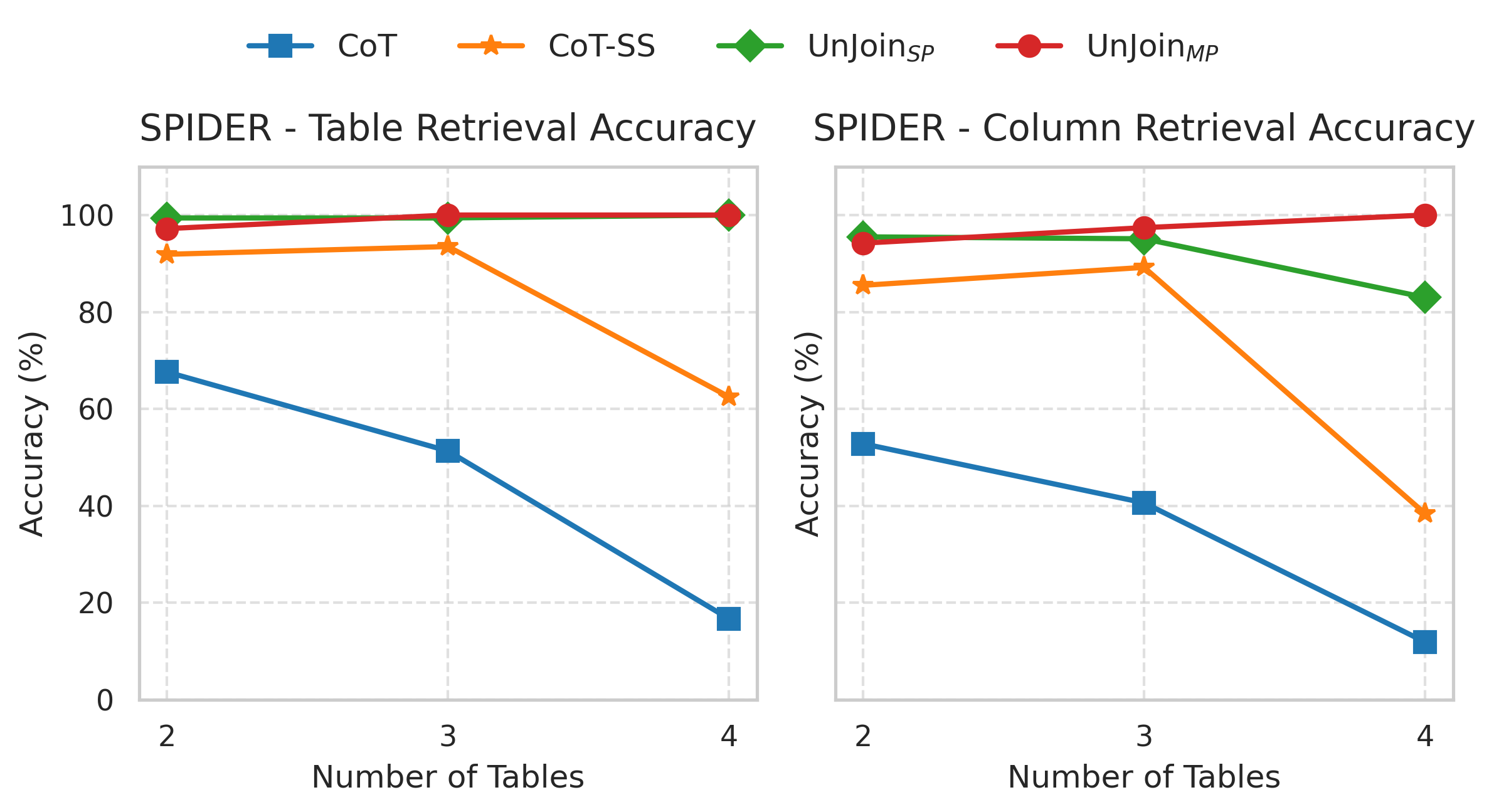}
    \vspace{-2.0em}
    \caption{Retrieval accuracy Performance with Increasing Number of Tables}
    \vspace{-0.75em}
         \label{fig:plot}
\end{figure}

\vspace{-0.5em}
\paragraph{How robust is \methodName with increasing relevant tables?} We analyze how table and column retrieval accuracy is affected as the number of  relevant tables needed to answer the query increases. As shown in Figure \ref{fig:plot}, baseline methods exhibit a sharp decline in performance as query complexity grows, whereas \methodName sustains high retrieval accuracy. This demonstrates the scalability and robustness of our approach in handling complex multi-table queries. Baselines often fail on multi-operation queries as the number of tables increases, while \methodName’s separate translation phase handles these systematically.

\begin{table*}[t!]
\small
\centering

\begin{tabular}{ll|ccc|ccc}
\hline
& & \multicolumn{3}{c|}{\textbf{Query Execution (QE)}} & \multicolumn{3}{c}{\textbf{Exact Match (EM)}} \\

\textbf{Dataset} & \textbf{Model}  & \textbf{CoT-SS.} & \textbf{\methodName$_{\text{SP}}$} & \textbf{\methodName$_{\text{MP}}$} 
 & \textbf{CoT-SS.} & \textbf{\methodName$_{\text{SP}}$} & \textbf{\methodName$_{\text{MP}}$} \\
\hline

\multirow{5}{*}{Spider} & GPT-4o           & 14.59 & \textbf{96.35} & 94.89 & 11.68 & \textbf{76.13} & 76.00 \\
& Gemini 1.5       & 81.39 & 94.89 & \textbf{95.99} & 59.23 & 73.36 & \textbf{75.57} \\
& Llama 3.3 (70B)  & 91.91 & 94.49 & \textbf{99.90} & 58.05 & 69.62 & \textbf{77.13} \\

\cdashline{2-8}

& SQLCoder (34B)   & 05.12 & \textbf{83.75} & 51.74 & 04.03 & \textbf{53.05} & 25.36 \\

\cdashline{2-8}

& SQLCoder (7B)    & 16.48 & 26.00 & \textbf{58.80} & 11.83 & 06.67 & \textbf{35.46} \\

\hline
\multirow{5}{*}{BIRD} & GPT-4o           & 16.86 & 88.55 & \textbf{89.75} & 11.24 & \textbf{51.74} & 50.36 \\
& Gemini 1.5       & 67.40 & 78.51 & \textbf{81.73} & 38.58 & 44.73 & \textbf{48.40} \\
& Llama 3.3 (70B)  & 82.61 & 91.97 & \textbf{96.28} & 49.40 & 52.10 & \textbf{56.35} \\

\cdashline{2-8}

& SQLCoder (34B)   & 11.24 & \textbf{64.75} & 31.75 & 08.03 & \textbf{28.55} & 19.70 \\
% \hline
\cdashline{2-8}
& SQLCoder (7B)    & 08.53 & 20.01 & \textbf{35.34} & 04.70 & 06.00 & \textbf{20.70} \\
\hline
\end{tabular}
\vspace{-0.75em}
\caption{QE and EM on \textbf{SPIDER} and \textbf{BIRD} datasets.}
\vspace{-1.5em}
\label{tab:spider_bird_qe_em}
\end{table*}

\vspace{-0.5em}
\paragraph{How does schema simplification benefit End2End table extraction?} From table \ref{tab:spider_bird_qe_em}, we see that CoT-SS suffers significant drops in
QE and EM, but from table \ref{tab:table_column_precision_recall_spider}, we can see that it shows strong table/column retrieval performance. This shows that schema simplification helps in retrieving correct table and column names, improving precision and recall. For instance, moving from CoT to  CoT-SS often increases recall (e.g., Llama 70B table recall on SPIDER goes from 0.41 to 0.74), showing that a simplified schema aids in finding relevant tables/columns.

\vspace{-0.75em}
\paragraph{Why does CoT-SS fails across all models?}
One other interesting thing to note is that the performance of CoT-SS on EM is generally poor across all the models. This highlights the importance of query translation step in our approach. Since CoT-SS directly generates the final SQL from the intermediate Simplified Schema, skipping the query translation step, it doesn't focus on integrating the correct compositional SQL operations, and thus, it omits necessary JOIN paths or introduce invalid syntax. This shows that query translation is essential in ensuring correct table linking and reducing structural inconsistencies.

\vspace{-0.75em}
\paragraph{How does our approach perform across different model sizes?}
From Table \ref{tab:spider_bird_qe_em}, we observe that our approach \methodName$_{\text{MP}}$ performs reasonably well with different model sizes. But our approach is most effective with larger LLMs due to the their superior instruction-following capability. One of the exceptions to the above trend is SQLCoder (34B), which, despite its size, does not perform well with \methodName. One reason can be that unlike general-purpose LLMs, SQLCoder is highly specialized for SQL generation and lacks strong natural language understanding and instruction-following capabilities, outside generating SQL. To gain a deeper understanding of \methodName's impact across different SQLCoder model sizes and its sensitivity to End2End extraction, please refer to Appendix~\ref{sec:discussion}.

\paragraph{Where do \methodName fails?} Detailed inspections of failure cases reveal three major error sources: \textbf{(a.) Misaligned Column Names:} Without the deterministic schema mapping, LLMs hallucinate or rename columns, leading to partial matches. 
\textbf{(b.)Ambiguity in Natural Language Queries:} Queries with unclear phrasing or multiple possible meanings often result in incorrect SQL generation. For instance, in the query "What are the names and release years for all the songs of the youngest singer?", the presence of similar column names (Song Name and Name) in the database creates ambiguity, leading to errors in column selection. \textbf{(c.) Unconventional Query Phrasing:} Users frequently use informal or unconventional phrasing in queries, making it challenging for the model to accurately interpret their intent. For example, in the query "For each singer name, what is the total sales for their songs?" the ordering condition needs adjustment to conform to proper SQL syntax.

\section{Comparison with Related Work}
\vspace{-0.5em}

\textbf{Seq2Seq methods.}Early Text-to-SQL systems like  Seq2SQL \cite{zhong2017seq2sqlgeneratingstructuredqueries} and SQLNet \cite{xu2017sqlnetgeneratingstructuredqueries} used Seq2Seq architectures but struggled with multi-table queries due to limited schema understanding. Schema-aware methods like IRNet \cite{guo-etal-2019-towards}, RAT-SQL \cite{wang2021ratsqlrelationawareschemaencoding}, and RESDSQL \cite{li2023resdsqldecouplingschemalinking}  improved performance by modeling schema relationships more explicitly. However, these methods often rely on manual schema linking and struggle with generalizing to unseen databases.

\textbf{PLMs based approaches.} Advancements in pre-trained transformer models (PLMs) like T5 \cite{raffel2023exploringlimitstransferlearning} and BART \cite{lewis2019bartdenoisingsequencetosequencepretraining} significantly boosted Text-to-SQL accuracy. Transformer-based models such as PICARD-T5 \cite{scholak2021picardparsingincrementallyconstrained}, TABSQLify \cite{nahid2024tabsqlifyenhancingreasoningcapabilities}, and BASE-SQL \cite{sheng2025basesqlpowerfulopensource} incorporated symbolic execution and constrained decoding to enhance multi-table reasoning. Still, their dependence on fine-tuning and pipeline complexity limits adaptability and scalability.

\textbf{LLMs prompts.} Prompt-based LLM methods like C3-SQL \cite{dong2023c3zeroshottexttosqlchatgpt}, DAIL-SQL \cite{gao2023texttosqlempoweredlargelanguage}, and MCS-SQL \cite{lee-etal-2025-mcs} bypass fine-tuning and use stepwise reasoning and prompt ensembling to improve performance. Yet, these models can be brittle, prompt-sensitive, and prone to producing invalid or incomplete SQL.
Other systems tackle schema complexity via query decomposition and retrieval (e.g., DATER \cite{ye2023largelanguagemodelsversatile}, ReActTable \cite{zhang2024reactable}, CHASE-SQL \cite{pourreza2024chasesqlmultipathreasoningpreference}), or modular pipelines like DIN-SQL \cite{pourreza2023dinsqldecomposedincontextlearning} and MAC-SQL \cite{wang2025macsqlmultiagentcollaborativeframework} that assign subtasks to dedicated agents. These approaches improve robustness, but add orchestration complexity, and are hard to generalize.

\textbf{SFT End2End methods.} SFT-based methods \cite{li2024codesbuildingopensourcelanguage, yang2024synthesizingtexttosqldataweak, pourreza-rafiei-2024-dts, talaei2024chesscontextualharnessingefficient, gorti2025mscsqlmultisamplecritiquingsmall, sheng2025basesqlpowerfulopensource} fine-tune open-source LLMs to improve SQL generation. Hybrid approaches like CHESS \cite{talaei2024chesscontextualharnessingefficient}, XiYan-SQL \cite{gao2025previewxiyansqlmultigeneratorensemble} and MSc-SQL \cite{gorti2025mscsqlmultisamplecritiquingsmall} combined prompt-based strategies with targeted fine-tuning, often leveraging smaller open-source models. Comprehensive surveys provide an overview of this evolving landscape \cite{qin2022surveytexttosqlparsingconcepts, katsogiannis-meimarakis-2023-survey, shi2024surveyemployinglargelanguage}.

Despite these advancements, challenges remain in generalization, scalability, and correctness for complex multi-table queries. Our framework, \methodName , addresses these gaps via schema simplification and decoupled query translation, achieving strong performance on Spider and BIRD.

\section{Conclusion and Future Work}
In this work, we propose \textbf{\methodName}, a two-step framework for multi-table Text-to-SQL that simplifies schemas for improved retrieval, generates an intermediate SQL query, and maps it back to the original schema. This modular approach reduces structural and retrieval errors, outperforming most of the baselines, including prompting-based, ICL and SFT based techniques, and reasoning models like DistilledDeepSeek Llama 70B. When layered over strong retrievers like ARM, JAR, Contriever, and DTR, \methodName significantly boosts end-to-end retrieval and SQL generation accuracy. By processing only schema information, \methodName ensures scalability and generalizability without pre-training or fine-tuning, advancing multi-table QA.

Future work includes extending \methodName to handle more complex data formats such as hierarchical, semi-structured, unstructured, and deeply nested schemas, enabling broader applicability across real-world databases and web tables.

\newpage
\section*{Limitations}
Our approach consistently outperforms existing baselines on BIRD and SPIDER while remaining scalable across various schemas and large databases. However, it has certain limitations. The effectiveness of \methodName depends on the LLM’s ability to accurately follow instructions, which may be less reliable with smaller or less capable models. Moreover, \methodName currently focuses on structural correctness rather than cell-level content extraction (e.g., resolving ambiguous entity names like “M. Obama” vs. “Michelle Obama”). As a result, it is less suited for tasks requiring precise entity disambiguation or record-level analysis. \methodName also does not work on multimodal/unstructured/semi-structured tables.

\section*{Ethics Statement}
We affirm that our work upholds the highest ethical standards in research and publication. Ethical considerations have been carefully addressed to ensure the responsible use of computational linguistics methodologies. To support reproducibility, we provide detailed information, including publicly available code, datasets, and relevant resources, all compliant with their respective ethical guidelines. Our claims are backed by experimental results, acknowledging minor variations due to the stochastic nature of black-box LLMs, which we mitigate by using a fixed temperature. Additionally, we outline dataset splits, model configurations, and prompting strategies to ensure transparency and reproducibility. AI was utilized in both experimentation and paper writing to enhance analysis, streamline result interpretation, and improve overall presentation clarity. Automated tools assisted in structuring content, refining language, and ensuring coherence, making the findings effectively communicated.

\section*{Acknowledgments}
We gratefully acknowledge the Cognitive Computation Group at the University of Pennsylvania and the Complex Data Analysis and Reasoning Lab at Arizona State University for their resources and computational support.

\bibliography{custom}

\newpage
\appendix
\section{Appendix: LLM Prompts and Examples}
\label{sec:prompts}
\subsection{Prompt for \methodName$_{\text{SP}}$}

\begin{lstlisting}[language=Prompt,caption={SQL Query Generation Prompt for \methodName$_{\text{SP}}$}]
You are an expert at semantic parsing. You have to follow two steps in order to complete your task.

Step 1: Getting the simplified query
You will be given a simplified schema which has only one table and multiple columns and a question. 
Please return the sql query with the correct format and syntax pertaining to that question. 
Remember to focus on getting the correct column extraction, and where clauses. 
DO NOT do any join operations. Treat this as a single table. The resulting query is the simplified query.

Example Table: bank_data
Column Name                  Description
customer.customer_id         Unique identifier for each customer
customer.name                Name of the customer
customer.gender              Gender of the customer
account.account_id           Unique identifier for accounts
account.balance              Current balance of the account
loan.loan_id                 Unique identifier for loans
loan.amount                  Loan amount
loan.status                  Status of the loan (e.g., Approved/Rejected)

Question 1:
"Which customers have an account balance greater than 10,000?"
SQL Query:
SELECT customer.customer_id, customer.name, account.balance
FROM bank_data
WHERE account.balance > 10000;

Question 2:
"List all customers with an approved loan."
SQL Query:
SELECT customer.customer_id, customer.name, loan.loan_id, loan.amount
FROM bank_data
WHERE loan.status = 'Approved';

Question 3:
"How many male customers have a loan?"
SQL Query:
SELECT COUNT(*) AS male_customers_with_loans
FROM bank_data
WHERE customer.gender = 'M' AND loan.loan_id IS NOT NULL;

Step 2: Getting the final query
Once you get the simplified query, you will have the following:

- A question: A natural language description of the desired query result.
- A simplified schema: A virtual table with columns in the format table_name.column_name, 
  combining data from multiple original tables.
- A simplified query: A SQL query written against the simplified schema.
- An original schema: A set of multiple related tables where the actual data resides.

Your task: Translate the simplified query into a query compatible with the original schema. 
Ensure the translated query aligns with the intent described in the question.

Follow these steps for translation:

1. Understand the Core Objective from the Question:
   - Identify the goal of the query (e.g., aggregate data, filter specific rows, join information across tables).

2. Map Simplified Schema Columns to the Original Schema:
   - Identify how the columns in the simplified schema correspond to tables and columns in the original schema.

3. Construct Necessary Joins:
   - If the original schema splits data across multiple tables, determine the joins needed to recreate the relationships.

4. Translate Filters and Conditions:
   - Map WHERE clauses, conditions, and filters in the simplified query to the original schema.

5. Adapt Query Logic (Aggregation, Sorting, etc.):
   - Match aggregations, grouping, or ordering logic from the simplified query to the original schema.

6. Validate the Final Query Against the Question:
   - Review the final query to ensure it satisfies the question and produces the intended result.

< examples>

Key Considerations:
- Use the Question as a Guide:
  Align the query logic with the intent expressed in the question.

- Simplified Schema as a Mapping Tool:
  Treat the simplified schema as a bridge between the question and the original schema.

- Validation:
  Ensure the translated query runs against the original schema and produces the intended result.
\end{lstlisting}

\subsection{Prompt for \methodName$_{\text{MP}}$ : Step1 : Query Generation on Simplified Schema}

\begin{lstlisting}[language=Prompt,caption={SQL Query Generation Prompt: \methodName$_{\text{MP}}$ Step 1}]
You are an expert at semantic parsing. You will be given a schema which has only one table and multiple columns and a question. Please return the SQL query with the correct format and syntax pertaining to that question. Remember to focus on getting the correct column extraction and WHERE clauses. DO NOT perform any join operations. Treat this as a single table.

<examples> 

Instructions:
1. The query should strictly adhere to the schema provided.
2. Ensure correct SQL syntax with SELECT, FROM, WHERE, GROUP BY, and ORDER BY clauses as needed.
3. The output query must be structured, readable, and executable in a standard SQL database.

Output:
\end{lstlisting}

\subsection{Prompt for \methodName$_{\text{MP}}$ : Step2 : Query Translation}

\begin{lstlisting}[language=Prompt,caption={SQL Query Generation Prompt: \methodName$_{\text{MP}}$ Step2}]
You are an expert at semantic parsing. You will be provided:

- A question: A natural language description of the desired query result.
- A simplified schema: A virtual table with columns in the format table_name.column_name, combining data from multiple original tables.
- A simplified query: A SQL query written against the simplified schema.
- An original schema: A set of multiple related tables where the actual data resides.

Your task:

1. Generate a SQL query based on a simplified schema.
2. Translate the simplified query into a query compatible with the original schema.
3. Ensure the translated query aligns with the intent described in the question.

### Steps for Translation:

1. **Understand the Core Objective from the Question**:
   - Identify the goal of the query (e.g., aggregate data, filter specific rows, join information across tables).

2. **Map Simplified Schema Columns to the Original Schema**:
   - Identify how the columns in the simplified schema correspond to tables and columns in the original schema.

3. **Construct Necessary Joins**:
   - If the original schema splits data across multiple tables, determine the joins needed to recreate the relationships.

4. **Translate Filters and Conditions**:
   - Map WHERE clauses, conditions, and filters in the simplified query to the original schema.

5. **Adapt Query Logic (Aggregation, Sorting, etc.)**:
   - Match aggregations, grouping, or ordering logic from the simplified query to the original schema.

6. **Validate the Final Query Against the Question**:
   - Review the final query to ensure it satisfies the question and produces the intended result.

<examples>

### Key Considerations:

- **Use the Question as a Guide**:
  - Align the query logic with the intent expressed in the question.
  - The question may highlight details (e.g., time ranges, specific groups) that must be included in the query.

- **Simplified Schema as a Mapping Tool**:
  - Treat the simplified schema as a bridge between the question and the original schema.
  - Focus on accurately mapping the simplified columns to the original schema.

- **Validation**:
  - Ensure the translated query runs against the original schema and produces the intended result.

**Output:**
\end{lstlisting}

\section{Schema Simplification Algorithm and Example}
\label{sec:schema_simplification}
\begin{algorithm}
\caption{Schema Simplification}
\begin{algorithmic}[1]
\State \textbf{Input:} Set of databases $D$
\State \textbf{Output:} Simplified schema dictionary $S$
\State $S \gets \{\}$ \Comment{Initialize empty dictionary}
\ForAll{database $d$ in $D$}
    \State $S[d.\text{name}] \gets [\,]$ \Comment{Initialize list for each database}
    \ForAll{table $t$ in $d.\text{tables}$}
        \ForAll{column name $c$ in $t.\text{columns}$}
            \State $s \gets t.\text{name} + \texttt{'.'} + c$
            \State Append $s$ to $S[d.\text{name}]$
        \EndFor
    \EndFor
\EndFor
\State \Return $S$
\end{algorithmic}
\end{algorithm}

\begin{figure*}
    \centering
    \includegraphics[width=1\linewidth]{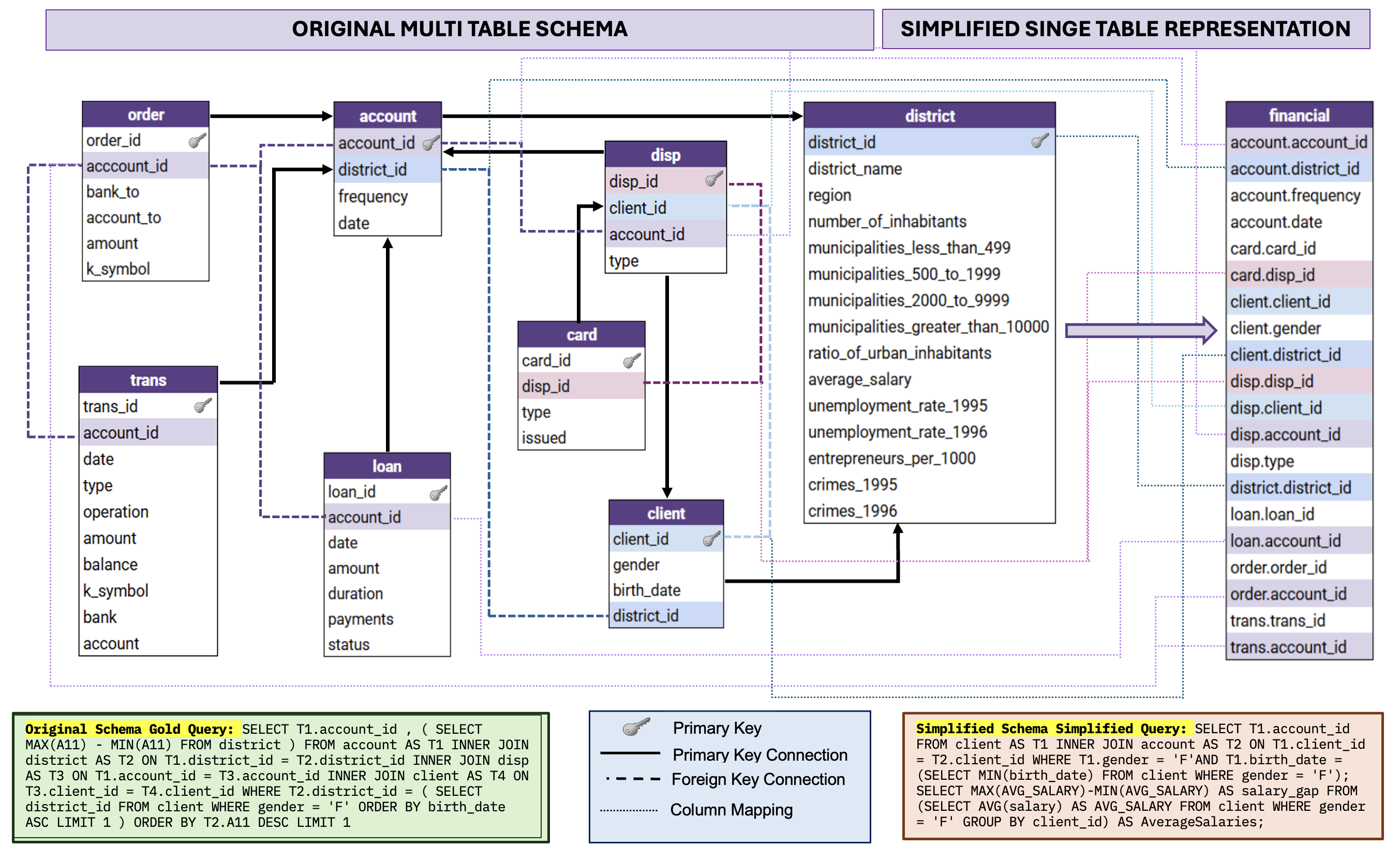}
    \caption{Schema Simplification}
    \label{fig:schema-simplification}
\end{figure*}

\section{Appendix: Baseline Results}

\begin{table}[h]
\small
\centering
% \footnotesize
% \renewcommand{\arraystretch}{0.9}
% \resizebox{\linewidth}{!}{%
\setlength{\tabcolsep}{4pt} % Adjust column spacing
\begin{tabular}{l|cc|cc|cc}
\hline
\addlinespace[3pt]
\multicolumn{7}{c}{\textbf{BIRD Dataset}} \\
\hline
\addlinespace[2pt]
& \multicolumn{2}{c|}{\textbf{GPT-4o}} & \multicolumn{2}{c|}{\textbf{Gemini}} & \multicolumn{2}{c}{\textbf{LLAMA 70B}} \\
\textbf{Method} & \textbf{QE} & \textbf{EM} & \textbf{QE} & \textbf{EM} & \textbf{QE} & \textbf{EM} \\
\hline
\addlinespace[2pt]
\multicolumn{7}{c}{\textit{Standard Prompting Baselines}} \\
\hline
\addlinespace[2pt]
CoT & 87.95  & 44.38 & 73.09 & 42.40 & 91.27 & 53.55 \\
PoT & 72.49 & 43.49 & 71.76 & 46.67 & 83.09 & 51.47 \\
MP       & 73.19 & 44.54 & 71.50 & 44.78 & 82.89 & 51.92 \\
SC    & 73.79 & 42.86 & 70.95 & 45.19 & 86.22 & 52.06 \\
\hline
\addlinespace[2pt]
\multicolumn{7}{c}{\textit{Table QA (SQL-based) baselines}} \\
\hline
\addlinespace[2pt]
TabSQLify           & 66.83 & 42.41 & 70.32 & 44.12 & 77.78 & 49.94 \\
\hline
\addlinespace[2pt]
\multicolumn{7}{c}{\textit{\methodName}} \\
\hline
\addlinespace[2pt]
\methodName$_{\text{SP}}$ & 88.55 & \textbf{51.74} & 78.51 & 44.73 & 91.97 & 52.10 \\
\methodName$_{\text{MP}}$  & \textbf{89.75 }& 50.36 & \textbf{81.73} & \textbf{48.40} & \textbf{96.28} & \textbf{56.35} \\
\hline
\end{tabular}
% }
\vspace{-0.5em}
\caption{QE and EM scores on BIRD dataset.}
\vspace{-1.0em}
\label{tab:bird_baselines}
\end{table}

\newpage
\section{Evaluation Metrics across different models}
\label{sec:PR_BIRD}

\begin{table*}[h]
\centering
\scriptsize  
\renewcommand{\arraystretch}{1.0}  
\setlength{\tabcolsep}{4pt}       
\resizebox{\textwidth}{!}{%
\begin{tabular}{l|cccc|cccc}
\hline
\textbf{Model} 
& \multicolumn{4}{c|}{\textbf{Table Precision}} 
& \multicolumn{4}{c}{\textbf{Table Recall}} \\
\cline{2-9}
& \textbf{CoT} & \textbf{CoT-SS} & \textbf{\methodName$_{\text{SP}}$} & \textbf{\methodName$_{\text{MP}}$} 
& \textbf{CoT} & \textbf{CoT-SS} & \textbf{\methodName$_{\text{SP}}$} & \textbf{\methodName$_{\text{MP}}$} \\
\hline
GPT-4o          & 59.30 & 57.20 & 73.60 & \textbf{74.70} & 58.60 & 58.00 & 73.40 & \textbf{74.20} \\
GPT-4o Mini     & 57.40 & 58.80 & \textbf{75.00} & 74.80 & 56.00 & 58.30 & 72.40 & \textbf{73.30} \\
Gemini 1.5 Flash & 53.00 & 62.90 & 65.30 & \textbf{66.20} & 52.40 & 61.00 & 65.30 & \textbf{65.70} \\
Mixtral         & 36.00 & 62.10 & 57.00 & \textbf{66.00} & 34.30 & 60.80 & 60.00 & \textbf{67.00} \\
SQLCoder-7B     & 56.00 & \textbf{62.60} & 22.10 & 60.60 & 50.70 & \textbf{63.10} & 23.30 & 58.60 \\
SQLCoder-34B    & 51.80 & 51.60 & \textbf{70.10} & 60.80 & 47.80 & 47.80 & \textbf{71.60} & 60.30 \\
Llama 3.1 (3B)  & \textbf{53.30} & 53.80 & 53.00 & 50.00 & \textbf{52.30} & 51.90 & 62.00 & 49.00 \\
Llama 3.3 (70B) & 67.10 & 74.30 & 75.30 & \textbf{76.80} & 67.20 & 71.20 & 74.80 & \textbf{75.00} \\
\hline
& \multicolumn{4}{c|}{\textbf{Column Precision}} 
& \multicolumn{4}{c}{\textbf{Column Recall}} \\
\hline
GPT-4o          & 68.80 & 68.00 & 86.30 & \textbf{86.60} & 68.20 & 65.80 & \textbf{85.40} & 85.00 \\
GPT-4o Mini     & 67.20 & 69.40 & 83.10 & \textbf{85.00} & 66.00 & 65.10 & 80.00 & \textbf{83.00} \\
Gemini 1.5 Flash & 54.40 & \textbf{88.60} & 82.80 & 83.80 & 53.30 & \textbf{85.40} & 82.90 & 82.30 \\
Mixtral         & 42.00 & 71.50 & 62.20 & \textbf{75.00} & 40.00 & 54.00 & 62.20 & \textbf{73.00} \\
SQLCoder-7B     & 55.60 & \textbf{69.80} & 17.10 & 62.50 & 52.60 & \textbf{66.40} & 20.20 & 61.10 \\
SQLCoder-34B    & 55.40 & 56.10 & \textbf{75.50} & 67.20 & 51.80 & 44.60 & \textbf{76.80} & 63.90 \\
Llama 3.1 (3B)  & \textbf{59.40} & 53.70 & 43.40 & 52.00 & \textbf{58.00} & 47.00 & 66.70 & 54.30 \\
Llama 3.3 (70B) & 81.80 & 84.80 & 85.00 & \textbf{86.00} & 81.60 & 82.90 & \textbf{84.90} & \textbf{84.90} \\
\hline
\end{tabular}
} % end of resizebox
\caption{Evaluation of Table/Column Precision and Recall on \textbf{BIRD} dataset.}
\label{tab:table_column_precision_recall_bird}
\end{table*}

\section{Further Discussion}
\label{sec:discussion}
\paragraph{Reason: SQLCoder 34B and 7B Underperformance.} SQLCoder 34B’s best table recall in \methodName$_{\text{MP}}$ (60.3 on BIRD) is lower than the best performing version of smaller general models like Llama 3.1 (62.0 on BIRD) (see Table \ref{tab:table_column_precision_recall_bird}) despite its larger size. 
This suggests that SQLCoder struggles not just with instruction following, but also with schema reasoning, since it is primarily trained on SQL generation rather than structured multi-table reasoning.

\paragraph{End2End Tables Extraction Sensitivity.} Across models, the gap between the best and worst-performing baselines is more pronounced for column precision and recall than for table precision/recall.
For instance, in SPIDER, GPT-4o’s column recall (95.5 in \methodName$_{\text{SP}}$) is significantly higher than SQLCoder-7B (54.1 in \methodName$_{\text{SP}}$), whereas the table recall gap is smaller. Similarly, we can observe this behavior within the same model with different sizes. For example, on the SPIDER dataset in the \methodName$_{\text{SP}}$ baseline, Llama 3.3 (70B) achieves a table recall of 74.0, whereas Llama 3.1 (3B) scores 65.9, showing a difference of only 8.1. However, the column recall for Llama 70B is 89.6, while Llama 3B achieves only 62, resulting in a much larger gap of 27.6. This trend indicates that smaller models struggle significantly more with precise column selection than table selection, likely due to their weaker contextual understanding and reasoning capabilities.
\end{document}